\documentclass{article}
\usepackage{spconf,amsmath,graphicx}
\usepackage{amsfonts}
\usepackage{amsmath}
\usepackage{bm}
\usepackage[colorlinks,linkcolor=blue]{hyperref}
\usepackage{url}
\usepackage{subcaption}

\renewcommand\vec{\mathbf}

\title{A Two-Stream Meticulous Processing Network for Retinal Vessel Segmentation}
\name{Author}
\address{Author Affiliation(s)}
\begin{document}
%
\maketitle
\begin{abstract}
Vessel segmentation in fundus is a key diagnostic capability in ophthalmology, and there are various challenges remained in this essential task. Early approaches indicate that it is often difficult to obtain desirable segmentation performance on thin vessels and boundary areas due to the imbalance of vessel pixels with different thickness levels. In this paper, we propose a novel two-stream Meticulous-Processing Network (MP-Net) for tackling this problem. To pay more attention to the thin vessels and boundary areas, we firstly propose an efficient hierarchical model automatically stratifies the ground-truth masks into different thickness levels. Then a novel two-stream adversarial network is introduced to use the stratification results with a balanced loss function and an integration operation to achieve a better performance, especially in thin vessels and boundary areas detecting. Our model is proved to outperform state-of-the-art methods on DRIVE, STARE, and CHASE\_DB1 datasets.
\end{abstract}

\vspace{-4pt}

\section{Introduction}

Fundus image analysis serves as a key and non-invasive tool in the diagnosis and treatment of many ophthalmological and cardiovascular diseases. Additionally, with the developing of deep learning methods, many network architectures based on U-Net or adversarial procedures have been proposed to learn the end-to-end relations between an original image and a ground-truth binary mask manually labeled by experts. Maninis \cite{maninis2016deep} proposed Deep Retinal Image Understanding (DRIU) which fine-tuned VGGNet. During the progress of deep learning approaches, segmentation performance on thin vessels has become a great challenge and focus. Zhang et al. \cite{zhang2018deep} propose a U-Net architecture (ML-UNet) \cite{ronneberger2015u} for multi-label segmentation of thin and stem (thick) vessels. Yan et al. \cite{yan2018joint} propose a novel segment-level loss in addition to the pixel-level loss to train a U-Net architecture (JL-UNet), and report increased segmentation accuracy for thin vessels. Yet, the work of Zhang et al. \cite{zhang2018deep} and Yan et al. \cite{yan2018joint} which propose an essentially multi-label miscellaneous network, do not have an end-to-end network which dedicated for specific binary classification tasks focusing different types of features. Additionally, Gu et. al \cite{gu2019net} propose a context encoder network (CE-Net) to better extract the high-level information of the image,  while the CE-Net loses to focus on thin and boundary areas.

In this paper, we inspect the rationale behind this problem from a perspective of data balancing. The reason that ordinary neural networks did not obtain desirable segmentation performance on thin vessels and boundary areas is that vessel data are suffered from imbalance internal to an assumed identical class (vascular or non-vascular). Vessels with different thickness levels may have different features for identification and localization, making them essentially different classes in a segmentation task. Therefore, balancing across these classes becomes an important work to avoid bias in learning. However, such balancing remains challenging as in most available segmentation datasets, the ground-truth mask is binary, providing no immediate information regarding thickness levels. In view of this challenge, we propose a novel morphological model that automatically segments and classifies (stratifies) ground-truth masks into strata regarding vessel thickness levels using hierarchical opening operations. In order to further increase the segmentation performance, we also propose a two-stream model that learns both general retinal vascular features and those specific to thin vessels and boundary areas by processing both all strata and only the thin vessels (the following "thin vessels" refer to both thin vessels and boundary areas) stratum. The results from the two streams are united (pixel-wise ORed) to output the final result.

Our contributions mainly lie in 3 aspects. (1) We propose a novel two-stream architecture to synthesize features of different thickness levels. (2) An efficient hierarchical model of opening operations, which automatically stratifies the ground-truth masks to inject thickness levels sensitivity to our model and is jointly utilized with a proposed CE-GAN model whose generator is based on the CE-Net \cite{gu2019net} architecture. (3) A balanced loss function and an integration operation to unify and enable weighing on vessel classes of various thickness levels. 

\begin{figure*}
	\includegraphics[width=\linewidth]{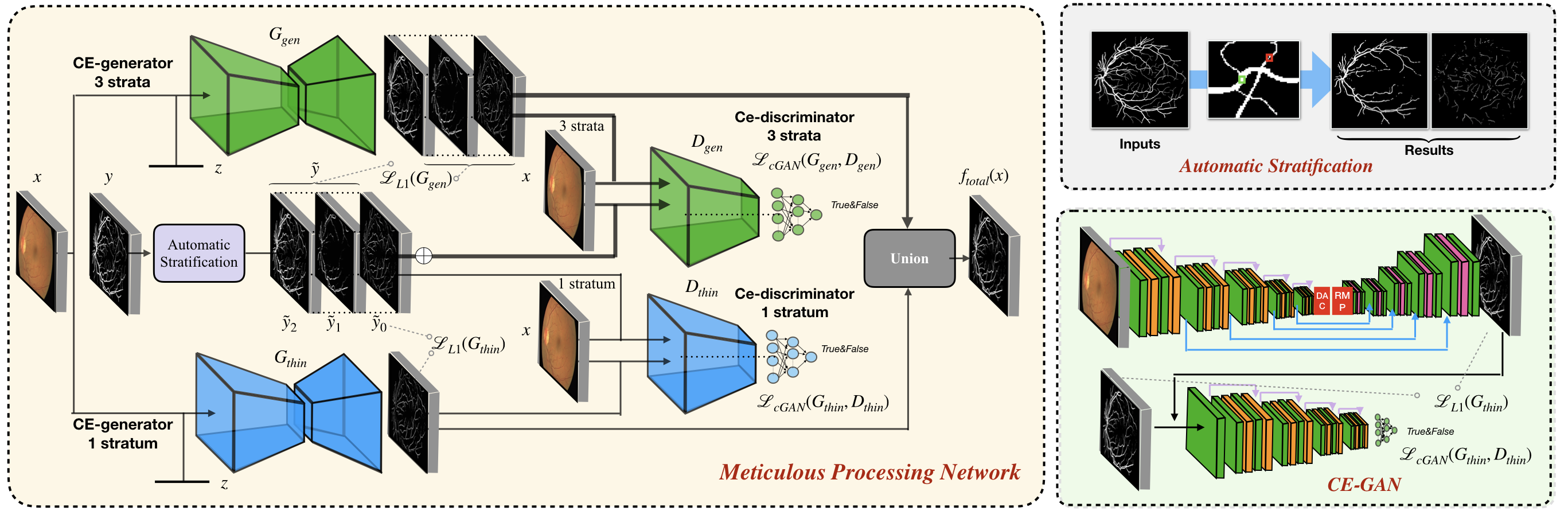}
	\caption{The proposed method includes a meticulous processing architecture, an automatic stratification method and CE-GAN network, where CE-GAN and automatic stratification are embedded in the meticulous processing architecture.}
	\centering
	\vspace{-10pt}
\end{figure*}
\vspace{-4pt}

\section{Proposed Method}

\subsection{Automatic Stratification}

For each original sample $(\vec{x}, \vec{y})$, the mask $\vec{y}$ is stratified into $n$ componential masks (strata): $\tilde{\vec{y}}_c, c=0,1,...,n-1$, each with only the vessel labels of the corresponding thickness levels. The stratification is achieved via opening (erosion then dilation) with thresholding kernels. For the opening operation, we apply thresholds for kernels sizes: $d_i\times d_i$, $i=1,2,...,n-1$. We define the diameter of the vessel as the discrete Fr\'{e}chet distance between its two border curves $\vec{A}$ and $\vec{B}$:

\begin{equation}
	\delta = \inf_{\alpha, \beta} \max_{t\in [0,1]} \{\delta_C(\vec{A}(\alpha(t)), \vec{B}(\beta(t)))\},
\end{equation}
where $\alpha$ and $\beta$: $[0, 1]\rightarrow [0, 1]$ are two non-decreasing surjections and $\delta_C(\cdot, \cdot)$ is the Chebyshev distance between two pixels. All vessels of $\delta\leq d$ are guaranteed to be completely erased via a $(d+1)\times (d+1)$ kernel, while all vessels with $\delta>d$ (attenuated during erosion) restore their original outlines after dilation and are intact from the whole opening process. This process results in an intermediary semi-limited mask $\vec{M}_{\delta>d}$, wherefrom we can derive the final precisely selective strata:
\begin{equation}
	\tilde{\vec{y}}_c := \vec{M}_{d_c<\delta\leq d_{c+1}} = \vec{M}_{\delta>d_c} - \vec{M}_{\delta>d_{c+1}},
\end{equation}
where $\vec{M}_{\delta>d_0}:=\vec{y}, \vec{M}_{\delta>d_n}:=\vec{0}$.

\subsection{Two-stream Model}

In order to learn vessel features of different specificities, we propose a novel two-stream model for both general features and those especially related to thin vessels. On one stream, $N_g$ learns general features via training against 3 strata. To effectively learn the features of different thickness levels, we propose to concatenate both the two stratified masks $\tilde{\vec{y}}_c$ (stem and thin) and the original mask $\vec{y}$ (raw) along a third, strata dimension to form $\tilde{\vec{y}}$ of shape $3\times h\times w$ for later training. Samples with stratified masks $(\vec{x}, \tilde{\vec{y}})$ are fed to a general end-to-end U-Net-like segmentation network $N_g$ that outputs a prediction map against each strata. On the other stream, an additional end-to-end network $N_t$ dedicated for segmenting thin vessels outputs only one prediction map against only the stratum of thin vessels labels $\tilde{\vec{y}}_0$.

We use weighted MSE as the losses of the network and apply corresponding backward updates to it. In this way, vessels of different thickness levels have configurable weights in the final losses and the thickness-insensitive segmentation dataset are able to be internally balanced:

\begin{equation}
	\begin{cases}
		\mathcal{L}_{gen} = \sum_{c=0}^2 w_c||N_g(\vec{x})_c - \tilde{\vec{y}}_c||_F	 \\
		\mathcal{L}_{thin} = ||N_t(\vec{x}) - \tilde{\vec{y}}_0||_F
	\end{cases},
\end{equation}
where $||\cdot||_F$ stands for the Frobenius norm of the residual tensor. 

The segmentation problem can also be formulated as an image-to-image translation task from the original image to the ground-truth mask. Specifically, we materialize the two-stream network as adversarial CE-GAN models. Under this context, we train the generative networks from those following loses:
\begin{equation}
	\begin{cases}
		\begin{aligned}
		\mathcal{L}_{cGAN}(G_g,D_g) & = \mathbb{E}_{\vec{x},\tilde{\vec{y}}}[\log D_g(\vec{x},\tilde{\vec{y}})] \\ & + \mathbb{E}_{\vec{x},\vec{z}}[\log (1-D_g(\vec{x},G_g(\vec{x},\vec{z})))]
		\end{aligned} \\
		\begin{aligned}
		\mathcal{L}_{cGAN}(G_t,D_t) & = \mathbb{E}_{\vec{x},\tilde{\vec{y}}_0}[\log D_t(\vec{x},\tilde{\vec{y}}_0)] \\ & + \mathbb{E}_{\vec{x},\vec{z}}[\log (1-D_t(\vec{x},G_t(\vec{x},\vec{z})))]
		\end{aligned}
	\end{cases}
\end{equation}

In addition, generators are also trained directly against the ground-truth strata to refine the segmentation results with L1 norm $\mathcal{L}_{L1}(G_g)$ and $\mathcal{L}_{L1}(G_t)$. Moreover, the adversarial segmentation networks are updated using a min-max algorithm, where the losses of the above two training ends are regularized by a hyper-parameter $\lambda$:

\begin{equation}
	\begin{cases}
		G_g^*=\arg\min_{G_g}\max_{D_g}\mathcal{L}_{cGAN}(G_g,D_g)+\lambda\mathcal{L}_{L1}(G_g) \\
		G_t^*=\arg\min_{G_t}\max_{D_t}\mathcal{L}_{cGAN}(G_t,D_t)+\lambda\mathcal{L}_{L1}(G_t)
	\end{cases}
\end{equation}

\begin{table*}[!t]
\setlength{\tabcolsep}{1mm}
\centering
\small
\setlength{\tabcolsep}{1.8 mm}
\caption{Performance Comparisons with Previous Work}
\begin{tabular}{l|l|llll|llll|llll}
\hline
\multicolumn{2}{l}{} & \multicolumn{4}{c}{\textbf{DRIVE}} & \multicolumn{4}{c}{\textbf{STARE}} & \multicolumn{4}{c}{\textbf{CHASE\_DB1}} \\
\textbf{Methods} & \textbf{Year} & $\bm{Sens }$ & $\bm{Spec}$ & $\bm{Acc}$ & $\bm{AUC}$ & $\bm{Sens}$ & $\bm{Spec}$ & $\bm{Acc}$ & $\bm{AUC}$ & $\bm{Sens}$ & $\bm{Spec}$ & $\bm{Acc}$ & $\bm{AUC}$ \\ 
\hline
\multicolumn{10}{c}{Unsupervised} \\
\hline
Zhang \cite{zhang2016robust} & 2016 & 0.7743 & 0.9725 & 0.9476 & 0.9636 & 0.7791 & 0.9758 & 0.9554 & 0.9748 & 0.7626 & 0.9661 & 0.9452 & 0.9606 \\
Fan \cite{fan2018hierarchical} & 2019 & 0.736 & 0.981 & 0.960 & - & 0.791 & 0.970 & 0.957 & - & 0.657 & 0.973 & 0.951 & - \\
\hline
\multicolumn{10}{c}{Classical Supervised} \\
\hline
Fraz \cite{fraz2012ensemble} & 2012 & 0.7406 & 0.9807 & 0.9480 & 0.9747 & 0.7548 & 0.9763 & 0.9534 & 0.9768 & 0.7224 & 0.9711 & 0.9469 & 0.9712 \\
Wang \cite{wang2019blood} & 2019 & 0.7648 & 0.9817 & 0.9541 & - & 0.7523 & 0.9885 & 0.9603 & - & 0.7730 & 0.9792 & 0.9603 & - \\
\hline
\multicolumn{10}{c}{Deep Learning} \\
\hline
Maninis \cite{maninis2016deep} & 2016 & 0.8280 & 0.9728 & 0.9541 & 0.9801 & 0.7919 & 0.9827 & 0.9706 & 0.9814 & 0.7651 & 0.9822 & 0.9657 & 0.9746 \\
ML-UNet \cite{zhang2018deep} & 2018 & 0.8723 & 0.9618 & 0.9504 & 0.9799 & 0.7673 & 0.9901 & 0.9712 & 0.9882 & 0.7667 & 0.9825 & 0.9649 & 0.9839 \\
JL-UNet \cite{yan2018joint} & 2018 & 0.7653 & 0.9818 & 0.9542 & 0.9752 & 0.7581 & 0.9846 & 0.9612 & 0.9801 & 0.7633 & 0.9809 & 0.9610 & 0.9781 \\
Gu \cite{gu2019net} & 2019 & 0.8309 & - & 0.9545 & 0.9779 & - & - & - & - & - & - & - & - \\
\textbf{Proposed} & 2019 & 0.7862 & \textbf{0.9858} & \textbf{0.9681} & \textbf{0.9844} & \textbf{0.7934} & 0.9884 & \textbf{0.9733} & \textbf{0.9883} & 0.7492 & \textbf{0.9890} & \textbf{0.9722} & \textbf{0.9858}
\\ \hline
\end{tabular}
\centering

\vspace{-10pt}
\end{table*}

\begin{table}[!h]
	\centering
	\caption{$AUC$s of ablation study of the MP-Net}
	\small
	\setlength{\tabcolsep}{3 mm}
	\begin{tabular}{l|lll}
		\hline
		 & DRIVE & STARE & DB1 \\
		\hline
		CE-Net \cite{gu2019net} & 0.9779 & 0.9810 & 0.9806 \\
		CE-GAN & 0.9820 & 0.9817 & 0.9812 \\
		CE-GAN + stratify & 0.9839 & 0.9850 & 0.9840 \\
		CE-GAN + stratify + thin & \textbf{0.9844} & \textbf{0.9883} & \textbf{0.9858}
	\\ \hline
	\end{tabular}
	\centering
	
	\vspace{-10pt}
\end{table}
Since both the two networks produce smooth predictions, first we binarize the preliminary outputs with a threshold of 127. Then as the final outputs of our system, positive binarized predictions are united (pixel-wise ORed) with that from each prediction maps.

\section{Experiments}

\subsection{Datasets and Experimental Setup}

We evaluate our model on three standard datasets widely used for the retinal vessels segmentation task. All of these three datasets contain no annotations of vessels thickness levels and are therefore appropriate for our stratification model to process. DRIVE \cite{staal_ridge-based_2004} \footnote{\url{https://www.isi.uu.nl/Research/Databases/DRIVE/}} contains 40 color fundus (CF) images with manually labeled ground-truth masks, where 20 images for training and use the remaining 20 images for testing. To reduce selection bias, we repeat the experiment 5 times and report the averaged result. STARE \cite{hoover_locating_1998} \footnote{\url{http://cecas.clemson.edu/~ahoover/stare/}} contains 20 manually labeled CF images. We report average results on 4-fold cross-validation with 15 training samples and 5 testing samples. CHASE\_DB1 \cite{owen2009measuring} \footnote{\url{https://blogs.kingston.ac.uk/retinal/chasedb1/}} contains 28 labeled samples, where we report average performances on 4-fold cross-validation.

\subsection{Evaluation Metrics}

Standard metrics for binary classification tasks including Area Under Curve ($AUC$) of Receiver Operating Characteristic (ROC), Accuracy ($Acc$), Specificity ($Spec$), and Sensitivity ($Sens$) (Recall) are used for evaluating our model. The definitions of the selected metrics are given by: $Acc = \frac{TP+TN}{TP+TN+FP+FN}$, $Sens = \frac{TP}{TP+FN}$, and $Spec = \frac{TN}{TN+FP}$, where $TP$, $TN$, $FP$, and $FN$ respectively stand for true positives, true negatives, false positives, and false negatives.

\begin{figure}
	\begin{subfigure}{0.24\linewidth}
		\includegraphics[width=\linewidth]{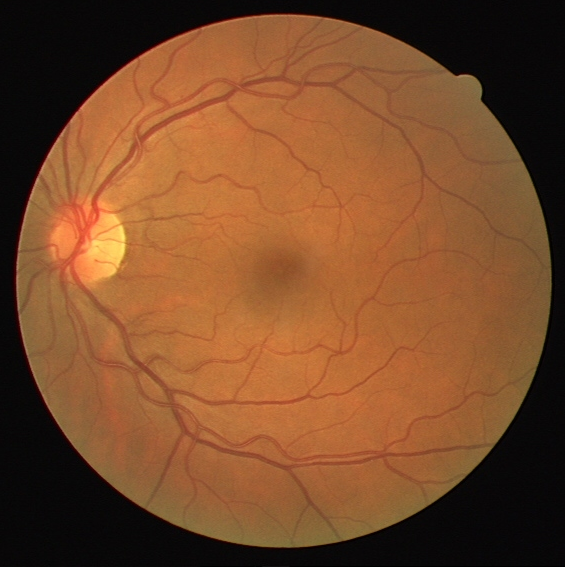}
	\end{subfigure}
	\begin{subfigure}{0.24\linewidth}
		\includegraphics[width=\linewidth]{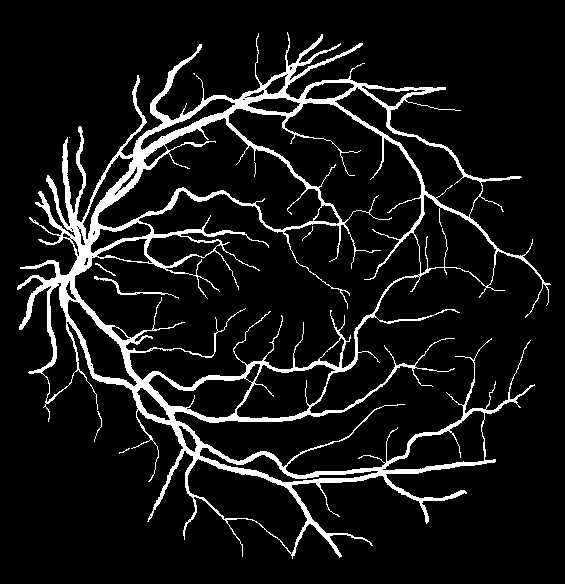}
	\end{subfigure}
	\begin{subfigure}{0.24\linewidth}
		\includegraphics[width=\linewidth]{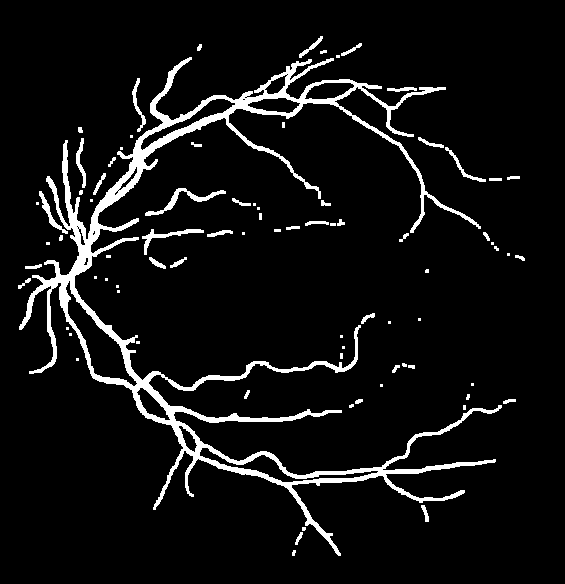}
	\end{subfigure}
	\begin{subfigure}{0.24\linewidth}
		\includegraphics[width=\linewidth]{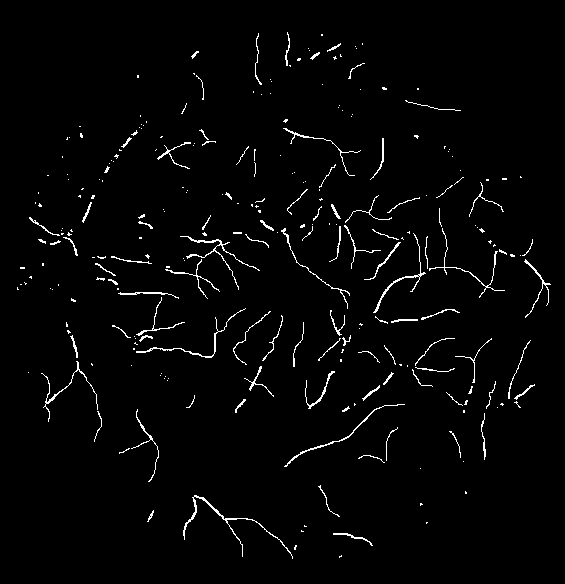}
	\end{subfigure}
	
	\begin{subfigure}{0.24\linewidth}
		\includegraphics[width=\linewidth]{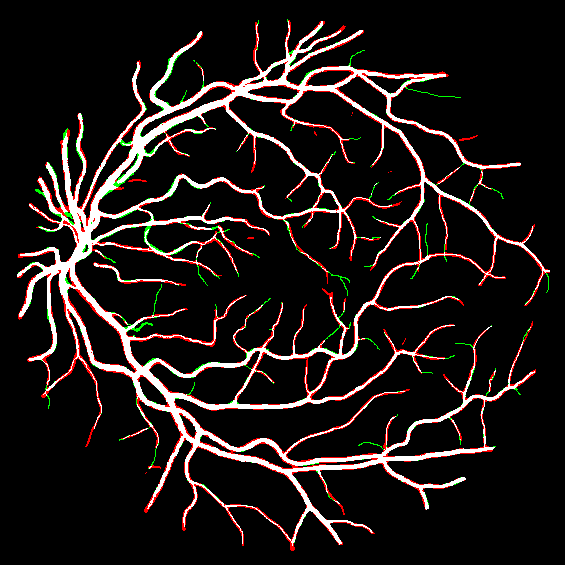}
	\end{subfigure}
	\begin{subfigure}{0.24\linewidth}
		\includegraphics[width=\linewidth]{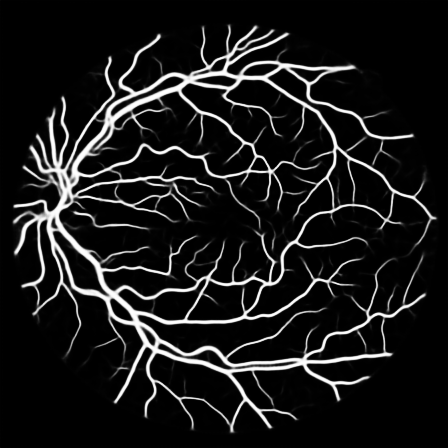}
	\end{subfigure}
	\begin{subfigure}{0.24\linewidth}
		\includegraphics[width=\linewidth]{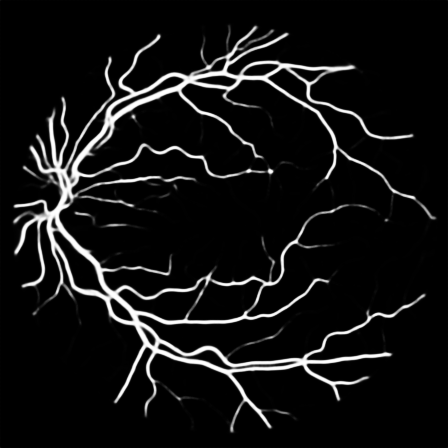}
	\end{subfigure}
	\begin{subfigure}{0.24\linewidth}
		\includegraphics[width=\linewidth]{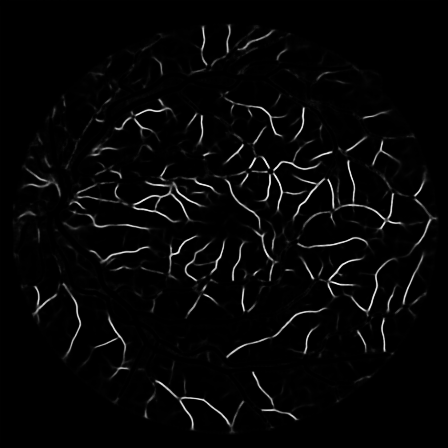}
	\end{subfigure}
	\caption{An example from the DRIVE dataset. Stratification (first row, left to right): (1) input image, (2) raw mask, (3) stem mask, (4) thin mask; Segmentation Results (second row): (5) overall prediction (red are false positive area while green are false negative area), (6) raw prediction, (7) stem prediction, (8) thin prediction (of the $N_g$ stream)}
	\vspace{-10pt}
\end{figure}

\subsection{Experimental Results}

To justify the performance of our model, we compare the 4 metrics with 8 representative previous works from all 3 open-access datasets. The comparison results presented in Table 1 show that our MP-Net model outperforms the state-of-the-art methods regarding accuracy and $AUC$ in all three datasets, which meter the practical prediction quality and the overall prediction quality independent on thresholding specifications. The $AUC$ advancement is greater in the DRIVE dataset. It's related to the fact that the DRIVE dataset contains more thin vessels, which is the main target of our model. Specificity is also the highest in DRIVE and CHASE\_DB1 while sensitivity is highest in STARE. Particularly, our method outperforms ML-UNet \cite{zhang2018deep} and JL-UNet \cite{yan2018joint} which adopt a different multi-class approach to also especially tackle the thin-vessels challenge. Figure 2 shows an example of our segmentation maps on DRIVE. As can be seen, most thin vessels and boundary areas have been meticulously picked up.

\subsection{Ablation Study}

Our proposed MP-Net can be roughly decomposed into 4 major progressive phases: (1) the backbone Context-Encoder Network (CE-Net) as a standalone generator segmenting non-stratified images, (2) the non-stratified CE-Net in (1) together with a discriminator to form a CE-GAN, (3) CE-GAN with a stratified CE-Net (i.e. with raw, stem, and thin strata) to form one stream of the MP-Net, and (4) The one-stream MP-Net in (3) with another stream of thin-stratum-specific GAN in (2) to form the complete two-stream MP-Net. We perform a whole series of ablation studies on all the datasets to verify the effect of each component via separation. The results in Table 2 validate that the stratification and mingled training mechanism and thin-specific designs are both effective improvements to the baseline system.

\section{Conclusion}

In this paper, we propose the Meticulous-Processing Network (MP-Net) which refines segmentation performance on thin vessels by stratifying and training on different thickness levels. The performance comparison and ablation study validate our design. This composited method can also be extended to more vessel-like segmentation tasks.

\bibliographystyle{IEEEbib}
\bibliography{mpnet}

\begin{thebibliography}{10}

\bibitem{maninis2016deep}
Kevis-Kokitsi Maninis, Jordi Pont-Tuset, Pablo Arbel{\'a}ez, and Luc Van~Gool,
\newblock ``Deep retinal image understanding,''
\newblock in {\em International conference on medical image computing and
  computer-assisted intervention}. Springer, 2016, pp. 140--148.

\bibitem{zhang2018deep}
Yishuo Zhang and Albert~CS Chung,
\newblock ``Deep supervision with additional labels for retinal vessel
  segmentation task,''
\newblock in {\em International Conference on Medical Image Computing and
  Computer-Assisted Intervention}. Springer, 2018, pp. 83--91.

\bibitem{ronneberger2015u}
Olaf Ronneberger, Philipp Fischer, and Thomas Brox,
\newblock ``U-net: Convolutional networks for biomedical image segmentation,''
\newblock in {\em International Conference on Medical image computing and
  computer-assisted intervention}. Springer, 2015, pp. 234--241.

\bibitem{yan2018joint}
Zengqiang Yan, Xin Yang, and Kwang-Ting Cheng,
\newblock ``Joint segment-level and pixel-wise losses for deep learning based
  retinal vessel segmentation,''
\newblock {\em IEEE Transactions on Biomedical Engineering}, vol. 65, no. 9,
  pp. 1912--1923, 2018.

\bibitem{gu2019net}
Zaiwang Gu, Jun Cheng, Huazhu Fu, Kang Zhou, Huaying Hao, Yitian Zhao, Tianyang
  Zhang, Shenghua Gao, and Jiang Liu,
\newblock ``Ce-net: Context encoder network for 2d medical image
  segmentation,''
\newblock {\em IEEE transactions on medical imaging}, 2019.

\bibitem{zhang2016robust}
Jiong Zhang, Behdad Dashtbozorg, Erik Bekkers, Josien~PW Pluim, Remco Duits,
  and Bart~M ter Haar~Romeny,
\newblock ``Robust retinal vessel segmentation via locally adaptive derivative
  frames in orientation scores,''
\newblock {\em IEEE transactions on medical imaging}, vol. 35, no. 12, pp.
  2631--2644, 2016.

\bibitem{fan2018hierarchical}
Zhun Fan, Jiewei Lu, Caimin Wei, Han Huang, Xinye Cai, and Xinjian Chen,
\newblock ``A hierarchical image matting model for blood vessel segmentation in
  fundus images,''
\newblock {\em IEEE Transactions on Image Processing}, vol. 28, no. 5, pp.
  2367--2377, 2018.

\bibitem{fraz2012ensemble}
Muhammad~Moazam Fraz, Paolo Remagnino, Andreas Hoppe, Bunyarit Uyyanonvara,
  Alicja~R Rudnicka, Christopher~G Owen, and Sarah~A Barman,
\newblock ``An ensemble classification-based approach applied to retinal blood
  vessel segmentation,''
\newblock {\em IEEE Transactions on Biomedical Engineering}, vol. 59, no. 9,
  pp. 2538--2548, 2012.

\bibitem{wang2019blood}
Xiaohong Wang, Xudong Jiang, and Jianfeng Ren,
\newblock ``Blood vessel segmentation from fundus image by a cascade
  classification framework,''
\newblock {\em Pattern Recognition}, vol. 88, pp. 331--341, 2019.

\bibitem{staal_ridge-based_2004}
Joes Staal, Michael~D Abr{\`a}moff, Meindert Niemeijer, Max~A Viergever, and
  Bram Van~Ginneken,
\newblock ``Ridge-based vessel segmentation in color images of the retina,''
\newblock {\em IEEE transactions on medical imaging}, vol. 23, no. 4, pp.
  501--509, 2004.

\bibitem{hoover_locating_1998}
Adam Hoover, Valentina Kouznetsova, and Michael Goldbaum,
\newblock ``Locating blood vessels in retinal images by piece-wise threshold
  probing of a matched filter response.,''
\newblock in {\em Proceedings of the {AMIA} {Symposium}}. 1998, p. 931,
  American Medical Informatics Association.

\bibitem{owen2009measuring}
Christopher~G Owen, Alicja~R Rudnicka, Robert Mullen, Sarah~A Barman, Dorothy
  Monekosso, Peter~H Whincup, Jeffrey Ng, and Carl Paterson,
\newblock ``Measuring retinal vessel tortuosity in 10-year-old children:
  validation of the computer-assisted image analysis of the retina (caiar)
  program,''
\newblock {\em Investigative ophthalmology \& visual science}, vol. 50, no. 5,
  pp. 2004--2010, 2009.

\end{thebibliography}

\end{document}